\documentclass[letterpaper, 10pt, conference]{ieeeconf}

\IEEEoverridecommandlockouts
\usepackage[T1]{fontenc}
\usepackage{times}

\title{FeelWorld: Visuo-Tactile World Model for Hierarchical Contact Prediction and Planning}
\author{
    Wenxuan Ma$^{1,2}$, Chaofan Zhang$^{1,2}$, Chao Xue$^{1,2}$, Yinghao Cai$^{1}$, Guocai Yao$^{3}$, \\ Shaowei Cui$^{1,2,3,*}$, and Shuo Wang$^{1}$ \vspace{1mm}\\
    $^{1}$Institute of Automation, Chinese Academy of Sciences \\
    $^{2}$Imprintx Robotics  \\
    $^{3}$Beijing Academy of Artificial Intelligence \\ 
    \vspace{-5mm}
    \thanks{$^{*}$Corresponding author: {\tt\small shaowei.cui@ia.ac.cn}
}
}

\usepackage[table]{xcolor}
\usepackage{amssymb}
\usepackage{amsmath,amsfonts}
\usepackage{mathtools}
\usepackage{algorithmic}
\usepackage{algorithm}
\usepackage{array}
\usepackage[strings]{underscore}
\usepackage{textcomp}
\usepackage{stfloats}
\usepackage{url}
\usepackage{verbatim}
\usepackage{graphicx}
\usepackage{cite}
\usepackage{hyperref}
\usepackage{siunitx}
\usepackage{booktabs}
\usepackage{multirow}
\usepackage{pifont}
\usepackage{arydshln}
\newcommand{\cmark}{\textcolor{green}{\checkmark}}
\newcommand{\xmark}{\textcolor{red}{$\times$}}
\begin{document}

\maketitle


\begin{abstract}
Humans plan physical interactions by imagining the possible outcomes of candidate actions. However, existing visual world models primarily capture appearance dynamics while overlooking the tactile states that govern contact-rich interactions, potentially producing imagined futures that appear visually plausible but violate physical dynamics. We introduce FeelWorld, a hierarchical visuo-tactile world model that jointly predicts future visual latents and three tactile states. FeelWorld organizes these states hierarchically as contact state, a 3D tactile latent that encodes force-related information, and slip state. These states are jointly predicted by a shared latent dynamics model with explicit supervision. To prevent irrelevant tactile signals during free-space motion from degrading visual prediction, we introduce a contact-gated asymmetric attention mechanism that maintains a visual-only prediction pathway before contact and enables joint visuo-tactile dynamics prediction during contact. The model is further trained with autoregressive rollouts and context noise injection to improve robustness to compounding errors. The predicted contact and slip states also support contact-aware CEM planning. Experiments on chip grasping, fruit grasping, and USB insertion show that FeelWorld reduces 10-step LPIPS from 0.084 to 0.058 and maintains an LPIPS that is 61\% lower than that of the visual baseline after an 80-step autoregressive rollout. FeelWorld also achieves an average zero-shot planning success rate of 81.7\%, providing an effective approach for incorporating tactile sensing into world models.
\end{abstract}

\section{Introduction}

Contact-rich manipulation requires robots to reason about physical interaction. Vision provides global scene context but offers only indirect evidence of contact physics. Tactile sensing directly measures force distribution, local deformation, and incipient slip, often capturing events that are occluded or produce negligible visual change~\cite{lee2020making,luo2025tactile}. These two modalities are complementary yet asymmetric: vision supplies continuous scene information, whereas tactile signals carry physical meaning only during contact.

Action-conditioned world models enable planning through imagined rollouts. The Dreamer family~\cite{hafner2020dreamer,hafner2025dreamerv3} learns compact latent dynamics for long-horizon imagination. Joint-embedding predictive architectures~\cite{assran2023self,assran2025v} avoid pixel reconstruction by predicting in a learned representation space. Recent visuo-tactile world models~\cite{lou2026dreamtac,tian2026vtwam,wu2026tactilewam} jointly predict visual and tactile futures, demonstrating that tactile grounding improves manipulation. However, these methods treat tactile signals as a monolithic prediction target without distinguishing the physical information encoded at different levels of interaction. Dense visual signals can overshadow sparse yet critical tactile information~\cite{ye2025dreamtacvla}. Without explicit physical structure, tactile signals provide weak supervision for latent dynamics. The open question is how to integrate tactile information hierarchically and selectively, respecting the physical structure of contact and fusing tactile signals only when they carry meaningful evidence.

We introduce FeelWorld, a hierarchical visuo-tactile world model. Our central observation is that tactile interaction decomposes into three hierarchical tactile states. \textit{Contact state} is binary and immediate, indicating whether the robot touches anything. The \textit{3D tactile latent} captures local surface deformation, encoding contact area, pressure distribution, and shear direction. \textit{Slip state} assesses interaction stability, an inherently temporal judgment. We predict these states in the order of contact state, 3D tactile latent, and slip state: the contact state controls visuo-tactile fusion, the 3D tactile latent represents local interaction geometry, and the slip state participates in the CEM planning cost.
\begin{figure}[t]
\centering
\includegraphics[width=\linewidth]{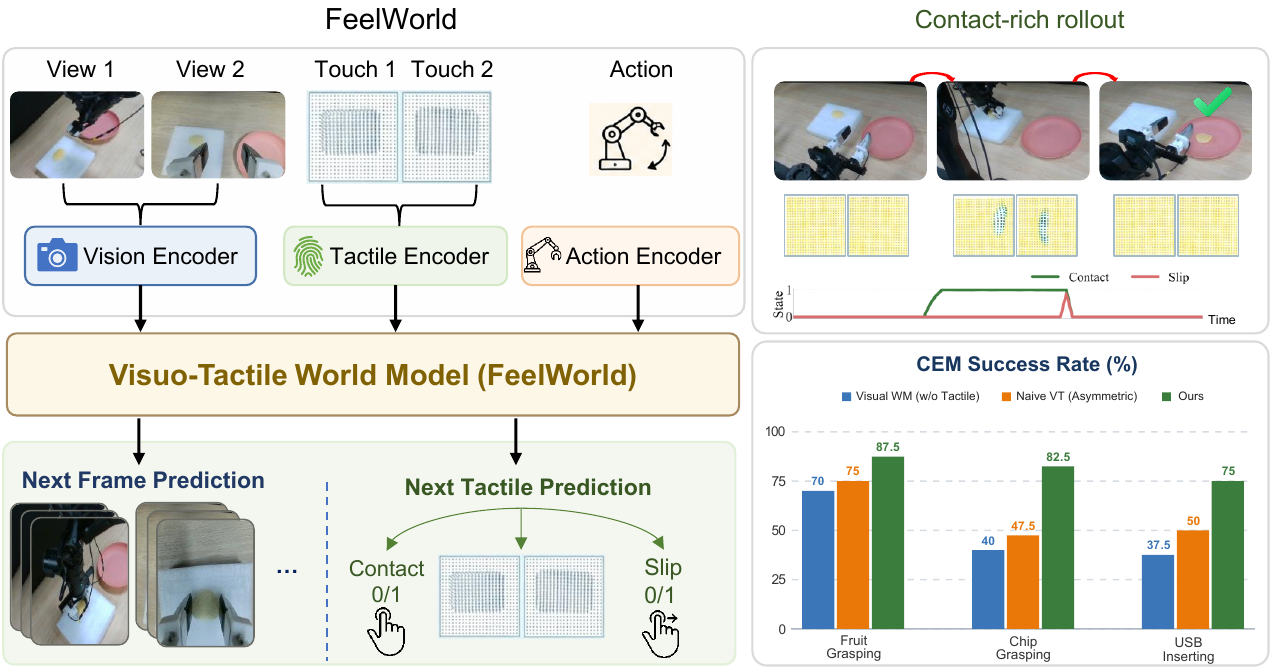}
\caption{\textbf{Overview of FeelWorld.} \textbf{Left}: multi-view visual observations, 3D tactile point clouds, and robot actions are encoded and fused to jointly predict future visual observations and hierarchical tactile states, including contact, 3D tactile latents, and slip. \textbf{Right}: contact-aware CEM planning consistently outperforms vision-only and naive visuo-tactile baselines across all tasks.}
\label{fig:intro}
\end{figure}

A second observation motivates our fusion design: tactile signals during free-space motion are sensor noise. We design a contact-gated asymmetric attention mechanism. Visual tokens retain a visual-only self-attention pathway, preserving an appearance-driven prediction path. Tactile tokens attend to visual tokens, grounding local contact geometry in global scene context. Predicted contact probability gates visuo-tactile cross-attention (visual tokens query tactile features), suppressing modulation when contact is absent and enabling it during contact.

We train with a joint objective combining visual and tactile latent prediction, explicit contact and slip supervision, and autoregressive rollout under context noise injection. At inference, a contact-aware Cross-Entropy Method (CEM) planner evaluates action sequences through imagined visuo-tactile rollouts, using predicted contact to transition from visual approach to tactile-grounded interaction. We evaluate on three contact-rich tasks: chip grasping, fruit grasping, and USB insertion.

Our contributions are:
\begin{itemize}
\item We propose \textbf{FeelWorld}, a hierarchical visuo-tactile world model that jointly predicts future visual latents and three tactile states, grounding imagined futures in contact physics. FeelWorld achieves an LPIPS of 0.058 and F1 scores of 98.1\% and 83.4\% for contact and slip prediction, respectively, while maintaining stable long-horizon rollouts.

\item We introduce a \textbf{Contact-gated Asymmetric Attention Mechanism}, which allows tactile features to modulate visual dynamics only during contact, thus protecting the world model from free-space sensor noise. Removing the gate degrades SSIM from 0.884 to 0.876 and PSNR from 28.89 to 27.52.

\item We develop a \textbf{Contact-aware CEM Planner}, which uses the predicted contact state to divide planning into a vision-guided stage and a joint visuo-tactile optimization stage, achieving an average success rate of 81.7\% across three contact-rich tasks and outperforming the vision-only baseline by 32.5 percentage points.
\end{itemize}

\section{Related Work}

\subsection{Action-Conditioned World Models}

World models learn action-conditioned dynamics, enabling robots to plan by imagining the outcomes of candidate actions. The Dreamer family supports long-horizon imagination through compact latent dynamics~\cite{hafner2020dreamer,hafner2025dreamerv3}. Recent robotic world models, such as V-JEPA 2~\cite{assran2025v}, Ctrl-World~\cite{guo2026ctrlworld}, and Interactive World Simulator~\cite{wang2026interactive}, further demonstrate action-conditioned visual imagination for planning, policy evaluation and improvement, and interactive simulation. Generative world models such as UniSim~\cite{yang2024unisim} and RoboDreamer~\cite{zhou2024robodreamer} extend this direction to interactive real-world simulation and compositional robot imagination. VT-WM~\cite{higuera2026vtwm} introduces touch into world models to improve the physical consistency and stability of long-horizon autoregressive rollouts. ContactWorld~\cite{zhang2026contactworld} highlights the value of spatially structured and temporally continuous visuo-tactile representations for contact-rich planning. However, most existing world models prioritize visual fidelity or treat touch as an undifferentiated observation, without explicitly modeling the structured contact dynamics underlying physical interaction. Consequently, their imagined futures may appear visually plausible yet remain physically inconsistent. FeelWorld addresses this limitation by jointly predicting visual latents and three tactile states with explicit hierarchical supervision and contact-gated fusion, enabling physically consistent prediction and planning.
\subsection{Visuo-Tactile Representation Learning}

Vision-based tactile sensors such as GelSight~\cite{yuan2017gelsight} and DIGIT~\cite{lambeta2020digit} capture contact geometry and force distribution through elastomer deformation, while 3D tactile point clouds provide geometric representations of surface interaction. Building on these sensors, self-supervised tactile pretraining~\cite{higuera2025sparsh}, self-supervised visuo-tactile alignment~\cite{kerr2023self}, and unified multimodal representation learning~\cite{fu2024touch,yang2024binding,cheng2025touch100k,feng2025anytouch} have enabled tactile representations that generalize across sensors and tasks. CLTP~\cite{ma2026cltp} further aligns 3D tactile point clouds with contact-aware semantics, while FG-CLTP~\cite{ma2026fg} introduces fine-grained semantic alignment for tactile representation pretraining. FeelWorld employs the pretrained FG-CLTP encoder to encode 3D tactile point clouds into latents that preserve force-related interaction information and jointly supervises three explicit hierarchical levels: contact state, 3D tactile latent, and slip state.
\subsection{Visuo-Tactile Manipulation Policies}

Prior visuo-tactile manipulation policies directly integrate tactile observations, force cues, or language-aligned tactile representations into action generation~\cite{huang20253dvitac,xue2025reactive,huang2026tafvla,zhang2026vtla,cheng2026omnivtla}.
More recent studies have begun to exploit predicted tactile or visuo-tactile futures. OmniVTA~\cite{zheng2026omnivta} combines short-horizon multimodal prediction with contact-aware control, while DreamTacVLA, Dream-Tac, VT-WAM, and Tactile-WAM use predicted multimodal observations or latent representations to support action generation~\cite{ye2025dreamtacvla,lou2026dreamtac,tian2026vtwam,wu2026tactilewam}. Despite these advances, most policy-oriented methods either map multimodal observations directly to actions or use future predictions mainly for policy generation, without explicitly structuring how contact states evolve under candidate actions. In contrast, our method integrates tactile sensing into an action-conditioned world model that captures the dynamics of contact-rich interactions, enabling physically consistent rollouts for future-state prediction and model-based planning.
\section{Method}

\subsection{Problem Formulation}
\label{subsec:problem_formulation}

FeelWorld learns an action-conditioned dynamics model that jointly predicts future visual latents and three tactile states. The visual input consists of images $I_t$ from $V$ views, such as third-person cameras for global scene context and a wrist camera for interaction feedback. A frozen visual encoder produces the visual latent $o_t=E_v(I_t)$.  A frozen tactile encoder produces the tactile latent $h_t=E_\tau(P_t)$. We denote the tactile states at time $t$ by the tuple $\tau_t=(c_t,h_t,s_t)$, comprising the contact state, 3D tactile latent, and slip state. Within the dynamics transformer, $z_t^o$ and $z_t^\tau$ denote the visual and tactile components, respectively, and $z_t=[z_t^o;z_t^\tau]$ denotes the fused visuo-tactile latent. The contact gate $g_t$ controls tactile-to-visual modulation (Sec.~\ref{subsec:contact_gated_attention}). Given robot proprioception $q_t$ and a 7-dimensional action $a_t$, the dynamics model predicts
\begin{equation}
    \hat{z}_{t+1}
    = [\hat{z}_{t+1}^o;\hat{z}_{t+1}^\tau]
    = \mathcal{W}_{\theta}(z_t,\, g_t,\, q_t,\, a_t).
    \label{eq:world_model}
\end{equation}
The visual component $\hat{z}_{t+1}^o$ is decoded into the next visual latent $\hat{o}_{t+1}$, while the tactile component $\hat{z}_{t+1}^\tau$ is decoded into the three tactile states $\hat{\tau}_{t+1}=(\hat{c}_{t+1},\hat{h}_{t+1},\hat{s}_{t+1})$, where $\hat{c}_{t+1},\hat{s}_{t+1}\in[0,1]$.

\begin{figure*}[t]
\centering
\includegraphics[width=\textwidth]{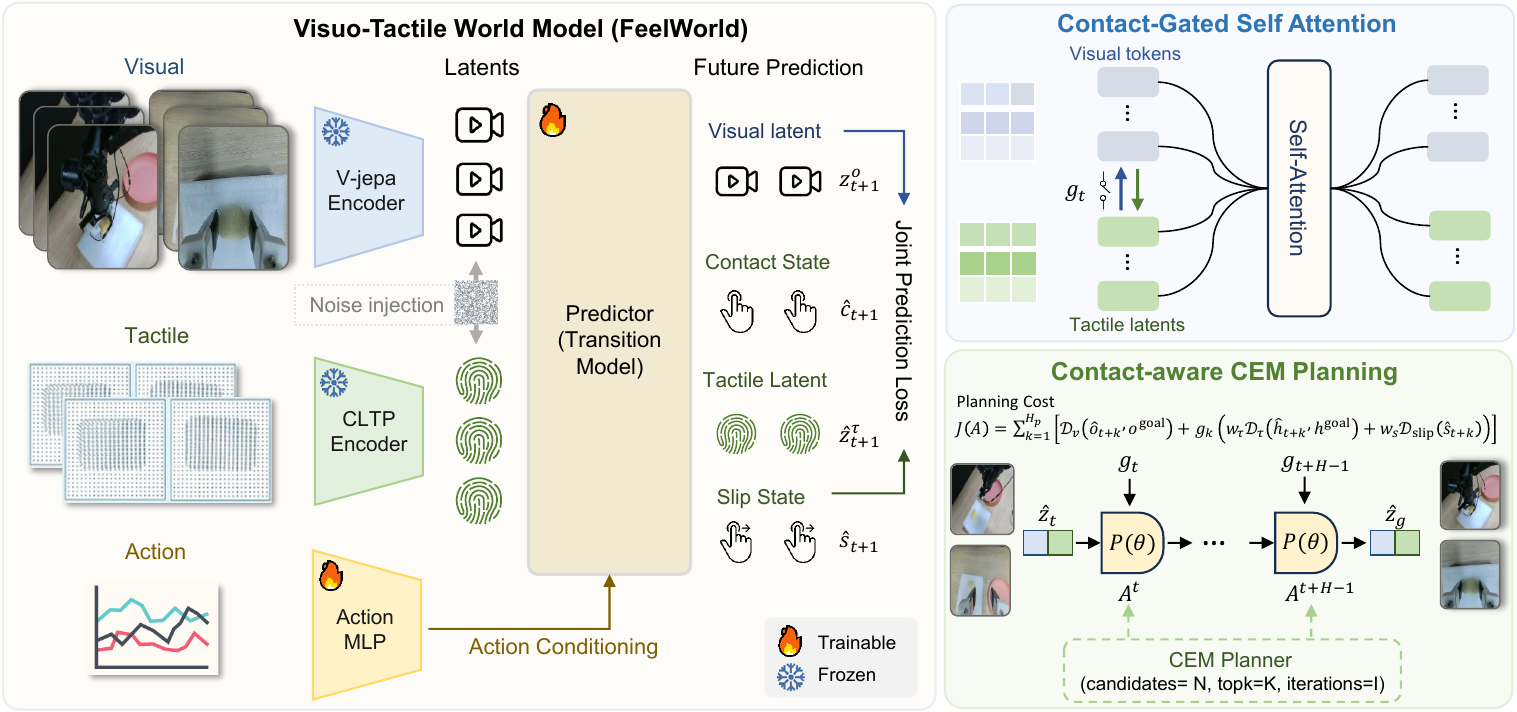}
\caption{\textbf{Architecture of FeelWorld.} Visual observations, tactile observations, and actions are encoded into latent tokens. Hierarchical tactile modeling decomposes touch into explicitly supervised contact states, 3D tactile latents, and slip states. Contact-gated asymmetric attention preserves visual prediction before contact and enables joint visuo-tactile prediction during contact. Autoregressive rollout training with context noise injection improves long-horizon robustness.}
\label{fig:method}
\end{figure*}

\subsection{Hierarchical Tactile Modeling}
\label{subsec:hierarchical_tactile}

We structure tactile prediction hierarchically around three states: contact state, 3D tactile latent, and slip state. Each state serves a distinct role in dynamics modeling and planning.

\textit{Tactile encoding.}
The 3D tactile point cloud $P_t$ is encoded by a pretrained encoder~\cite{ma2026fg} into a compact latent representation:
\begin{equation}
    h_t = E_\tau(P_t) \in \mathbb{R}^{d_\tau}.
    \label{eq:tactile_encoding}
\end{equation}

This latent encodes force-related contact information, including contact area,
pressure distribution, and principal deformation direction. We omit
LayerNorm for $h_t$, as normalization may remove magnitude information
associated with contact intensity.

\textit{Shared tactile prediction.}
The tactile prediction head decodes the next-step tactile latent from the predicted tactile component:
\begin{equation}
\hat{h}_{t+1}
=
H_{\tau}\bigl(\hat{z}_{t+1}^\tau\bigr),
\label{eq:tactile_dynamics}
\end{equation}
where $H_\tau$ is the tactile prediction head. This readout grounds the tactile component of the fused dynamics in the geometry and deformation measured by the tactile sensor.

\textit{Level 1: Contact state.}
A two-layer MLP contact head $H_c$ predicts binary contact probability from the predicted tactile component:
\begin{equation}
    \hat{c}_{t+1} = H_c\bigl(\hat{z}_{t+1}^\tau\bigr) \in [0,1].
    \label{eq:contact_pred}
\end{equation}
Contact is supervised with binary cross-entropy loss:
\begin{equation}
    \mathcal{L}_{\mathrm{contact}} = \mathrm{BCE}(\hat{c}_{t+1},\, c_{t+1}),
    \label{eq:loss_contact}
\end{equation}
where $c_{t+1}\in\{0,1\}$ is the ground-truth contact label. The predicted contact probability $\hat{c}_{t+1}$ controls visuo-tactile fusion (Sec.~\ref{subsec:contact_gated_attention}) and activates tactile objectives during planning (Sec.~\ref{subsec:contact_cem}).

\textit{Level 2: 3D tactile latent.}
The decoded 3D tactile latent $\hat{h}_{t+1}$ encodes local contact geometry and surface deformation. It is directly supervised against the stop-gradient target from the tactile encoder:
\begin{equation}
    \mathcal{L}_{\mathrm{tis}} = \|\hat{h}_{t+1} - \mathrm{sg}(E_\tau(P_{t+1}))\|_1,
    \label{eq:loss_tis}
\end{equation}
where $\mathrm{sg}(\cdot)$ denotes stop-gradient to prevent backpropagation through the frozen encoder. This $L_1$ loss grounds the predicted dynamics in physically measured deformation patterns.

\textit{Level 3: Slip state.}
Slip is inherently temporal; detecting incipient slip requires observing how the tactile component changes over time. A causal 1D convolutional head with kernel size $K{=}3$ aggregates a temporal window of predicted tactile features:
\begin{equation}
    \hat{s}_{t+1}
    = H_s\bigl(\hat{z}_{t-K+2:t+1}^\tau\bigr) \in [0,1],
    \label{eq:slip_head}
\end{equation}
where $H_s$ is the slip head. This architecture explicitly models the transition from the 3D tactile latent to the slip state. Slip labels are highly imbalanced (stable contact dominates), so we use binary focal loss~\cite{lin2017focal}:
\begin{equation}
    \mathcal{L}_{\mathrm{slip}} = -\alpha_t \bigl(1 - p_t\bigr)^\gamma \log p_t,
    \label{eq:loss_slip}
\end{equation}
where $p_t$ is the probability assigned to the ground-truth class and $\alpha_t$ is its class-balancing weight; we set $\alpha{=}0.95$ for positive (slip) examples and $\gamma{=}2.0$ to down-weight easy negatives. Because slip is defined only during contact, we compute $\mathcal{L}_{\mathrm{slip}}$ only at timesteps where $c_{t+1}{=}1$. The predicted slip probability $\hat{s}_{t+1}$ enters the planning cost to penalize unstable interactions.

\textit{Teacher-forcing objective.}
The visual prediction head decodes $\hat{o}_{t+1}=H_v(\hat{z}_{t+1}^o)$. Its next-state prediction is supervised against the stop-gradient target from the visual encoder $E_v$. Both $E_v$ and $E_\tau$ are pretrained and frozen, so only the dynamics predictor and prediction heads are trained:
\begin{equation}
    \mathcal{L}_{\mathrm{vis}}^{\mathrm{tf}}
    = \|\hat{o}_{t+1} - \mathrm{sg}(E_v(I_{t+1}))\|_1,
    \label{eq:loss_visual_tf}
\end{equation}
The joint teacher-forcing objective combines this visual loss with the hierarchical tactile losses:
\begin{equation}
    \mathcal{L}_{\mathrm{tf}}
    = \mathcal{L}_{\mathrm{vis}}^{\mathrm{tf}}
    + \lambda_\tau\mathcal{L}_{\mathrm{tis}}
    + \lambda_c\mathcal{L}_{\mathrm{contact}}
    + \lambda_s\mathcal{L}_{\mathrm{slip}},
    \label{eq:loss_tf}
\end{equation}
where $\lambda_\tau{=}0.3$, $\lambda_c{=}0.03$, and $\lambda_s{=}0.03$ balance the contribution of each tactile level.

\subsection{Contact-Gated Visuo-Tactile Attention}
\label{subsec:contact_gated_attention}

The encoded tactile latent $h_t$ and visual latent $o_t$ are projected into the tactile and visual transformer tokens $z_t^\tau$ and $z_t^o$, respectively. However, naive concatenation with standard self-attention allows noisy tactile signals during free-space motion to interfere with visual tokens, potentially degrading visual prediction. We therefore introduce contact-gated asymmetric attention. Tactile tokens first attend to visual tokens, grounding local tactile observations in global scene information, such as object pose, robot configuration, and task phase:
\begin{equation}
    \hat{z}_t^\tau
    = z_t^\tau
    + \mathrm{CrossAttn}(z_t^\tau,\,z_t^o,\,z_t^o).
    \label{eq:tactile_attend_visual}
\end{equation}

Conversely, whether visual tokens attend to tactile features is determined by the binary contact gate $g_t$, which is produced by the preceding contact prediction:
\begin{equation}
    \hat{z}_t^o
    = z_t^o
    + g_t\,\mathrm{CrossAttn}(z_t^o,\,\hat{z}_t^\tau,\,\hat{z}_t^\tau).
    \label{eq:visual_modulate}
\end{equation}
The complete fused feature is $\hat{z}_t=[\hat{z}_t^o;\hat{z}_t^\tau]$. Its visual component retains a visual prediction pathway while incorporating tactile evidence only during contact. At the initial timestep, the gate uses the ground-truth contact label from the first observation ($g_0 = c_0$). During teacher-forcing training, subsequent gates also use ground-truth contact labels ($g_t = c_t$), providing clean supervision for the fusion mechanism. During autoregressive rollout and inference, the gate is derived from the predicted contact probability ($g_t = \mathbb{I}[\hat{c}_t \geq 0.5]$). Visuo-tactile cross-attention is then disabled when $\hat{c}_t<0.5$ and enabled otherwise.

\subsection{Rollout Training with Context Noise Injection}
\label{subsec:rollout_noise}

The teacher-forcing objective of Eq.~\eqref{eq:loss_tf} trains the model to predict the next state from ground-truth context. At inference, however, the model operates autoregressively, receiving its own predicted latents as input. Prediction errors at each step become input errors at the next. This discrepancy causes errors to compound over long horizons, producing rollouts that diverge from physically plausible futures.

We mitigate this by training with multi-step autoregressive rollouts ($R$ steps) alongside the teacher-forcing objective. Starting from a randomly sampled timestep $t$, the model predicts forward for $R$ steps using its own fused outputs as context. Additionally, we inject small Gaussian noise into the visuo-tactile context before feeding it into the world model:

\begin{equation}
    \tilde{z}_{t+r} = \hat{z}_{t+r} + \varepsilon_z,
    \label{eq:noise_injection}
\end{equation}
with $\varepsilon_z \sim \mathcal{N}(0, \sigma^2 I)$. The noise scale $\sigma$ approximates the prediction error distribution encountered during inference-time autoregression. Training the model to predict from noisy contexts encourages it to generate high-quality outputs from perturbed inputs, thereby improving robustness during autoregressive rollout.

The rollout loss averages over all $R$ steps:
\begin{equation}
    \begin{aligned}
    \mathcal{L}_{\mathrm{rollout}}
    = \frac{1}{R}\sum_{r=1}^{R}\Bigl(
    &\mathcal{L}_{\mathrm{vis},r}^{\mathrm{roll}}
    + \lambda_\tau\mathcal{L}_{\mathrm{tis},r}^{\mathrm{roll}} \\
    &+ \lambda_c\mathcal{L}_{\mathrm{contact},r}^{\mathrm{roll}}
    + \lambda_s\mathcal{L}_{\mathrm{slip},r}^{\mathrm{roll}}\Bigr),
    \end{aligned}
    \label{eq:loss_rollout}
\end{equation}
where the four terms at each step $r$ denote the corresponding losses evaluated at $t{+}r$; $\mathcal{L}_{\mathrm{slip},r}^{\mathrm{roll}}$ is computed only when $c_{t+r}{=}1$. The total objective is $\mathcal{L}_{\mathrm{total}} = \mathcal{L}_{\mathrm{tf}} + \lambda_{\mathrm{roll}}\mathcal{L}_{\mathrm{rollout}}$.

\subsection{Contact-Aware CEM Planning}
\label{subsec:contact_cem}

At control step $t$, the current observation is encoded as $x_t=(z_t,g_t,q_t)$, where $z_t$ is the fused visuo-tactile latent. Given a candidate action sequence $A=(a_t,a_{t+1},\dots,a_{t+H_p-1})$, FeelWorld rolls $x_t$ forward for $H_p$ steps and predicts the fused latent together with its visual and tactile readouts, contact probability, and slip probability at every step. The planner can evaluate an action sequence in latent space before executing it on the robot.

We use CEM to search for the action sequence $A$ that minimizes the planning cost. The planner maintains a diagonal Gaussian distribution over action sequences. In each of $I$ iterations, it samples $N$ candidates, evaluates each candidate through an autoregressive model rollout, retains the $K$ sequences with the lowest cost, and refits the Gaussian to these elites. After the final iteration, the robot executes the first two actions of the lowest-cost sequence, observes the resulting state, and replans.

The planning cost retains visual task progress throughout the rollout and activates tactile interaction terms only after predicted contact:
{\small
\begin{equation}
\begin{multlined}
J(A) = \sum_{k=1}^{H_p}
\Bigl[ \mathcal{D}_{v}(\hat{o}_{t+k},\, o^{\mathrm{goal}}) \\
+ g_k\bigl(w_{\tau}\mathcal{D}_{\tau}(\hat{h}_{t+k},\, h^{\mathrm{goal}})
+ w_s\mathcal{D}_{\mathrm{slip}}(\hat{s}_{t+k})\bigr) \Bigr],
\end{multlined}
\label{eq:planning_objective}
\end{equation}
}
where $g_k = \mathbb{I}[\hat{c}_{t+k}\geq0.5]$ is the hard contact gate at future step $k$, and $w_\tau,w_s$ weight the tactile and slip costs. The visual distance $\mathcal{D}_v$ and tactile distance $\mathcal{D}_\tau$ are both $L_1$ distances in their respective latent spaces. $\mathcal{D}_{\mathrm{slip}}(\hat{s}_{t+k}) = \hat{s}_{t+k}$ directly penalizes predicted slip probability. The visual distance $\mathcal{D}_v$ remains active over the full horizon. Before predicted contact ($g_k=0$), tactile and slip costs are suppressed; after contact ($g_k=1$), the weighted tactile goal distance and slip penalty are added. Each candidate action sequence therefore induces its own contact-conditioned physical cost.
Algorithm~\ref{alg:contact_cem} summarizes one receding-horizon planning step.
\begin{algorithm}[t]
\caption{Contact-Aware CEM Planning}
\label{alg:contact_cem}
\footnotesize
\begin{algorithmic}[1]
\REQUIRE $\mathcal{W}_{\theta}$, $x_t$, $(o^{\mathrm{goal}},h^{\mathrm{goal}})$, $H_p$, $N$, $K$, $I$, $w_\tau$, $w_s$
\ENSURE $A_{t:t+1}^{\star}$
\STATE Initialize $q(A)=\mathcal{N}(\mu_0,\sigma_0^2 I)$
\FOR{$i=1$ to $I$}
    \STATE Sample $\{A^{(n)}\}_{n=1}^{N}\sim q(A)$
    \FOR{$n=1$ to $N$}
        \STATE $\{[\hat z_{t+k}^o;\hat z_{t+k}^\tau]\}_{k=1}^{H_p}\gets\mathcal{W}_{\theta}(x_t,A^{(n)})$
        \STATE Decode $\{\hat o_{t+k}\}_{k=1}^{H_p}$ from $\{\hat z_{t+k}^o\}_{k=1}^{H_p}$
        \STATE Decode $\{\hat h_{t+k},\hat c_{t+k},\hat s_{t+k}\}_{k=1}^{H_p}$ from $\{\hat z_{t+k}^\tau\}_{k=1}^{H_p}$
        \STATE $J_n\gets 0$
        \FOR{$k=1$ to $H_p$}
            \IF{$\hat c_{t+k}<0.5$}
                \STATE $J_n\gets J_n+\mathcal{D}_v(\hat o_{t+k},o^{\mathrm{goal}})$
            \ELSE
                \STATE $J_n\gets J_n+\mathcal{D}_v(\hat o_{t+k},o^{\mathrm{goal}})+w_\tau\mathcal{D}_\tau(\hat h_{t+k},h^{\mathrm{goal}})+w_s\mathcal{D}_{\mathrm{slip}}(\hat s_{t+k})$
            \ENDIF
        \ENDFOR
    \ENDFOR
    \STATE $\mathcal{E}\gets$ top-$K$ with lowest $J_n$
    \STATE $\mu,\sigma^2\gets\mathrm{Mean}(\{A^{(n)}\}_{n\in\mathcal{E}}),\mathrm{Var}(\{A^{(n)}\}_{n\in\mathcal{E}})$
\ENDFOR
\STATE \textbf{return} first two actions of $A^{(\arg\min_n J_n)}$
\end{algorithmic}
\end{algorithm}

Tactile goals are physically unreachable before contact. Applying $\mathcal{D}_\tau$ or $\mathcal{D}_{\mathrm{slip}}$ during free-space approach would make CEM optimize an unrealizable physical state and create competing objectives. The contact gate removes this conflict while preserving visual progress throughout the horizon: tactile geometry and slip influence planning only after contact becomes reachable.

\section{Experiments}

\begin{figure}[!t]
\centering
\includegraphics[width=0.7\linewidth]{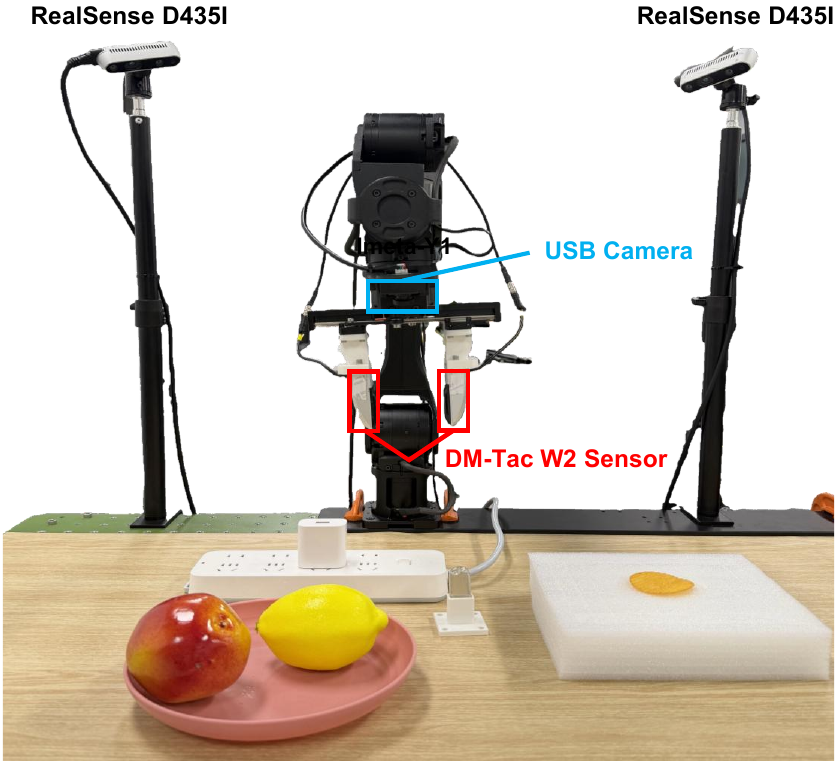}
\caption{\textbf{Robot platform.} The Imeta-Y1 robot is equipped with three RGB cameras and DM tactile sensors. FeelWorld uses the left external camera to observe the global state and the wrist-mounted camera to provide local observations. The DM tactile sensor captures 3D tactile point clouds and contact states.}
\label{fig:platform}
\end{figure}

We design experiments to answer three questions. Does joint visuo-tactile modeling with hierarchical tactile supervision improve visual prediction quality over vision-only baselines? Can FeelWorld capture the temporal dynamics of tactile interaction? Does contact-aware planning improve success rates in contact-rich manipulation compared with vision-only and naive joint planning baselines?
\subsection{Experimental Setup}
\subsubsection{Robot Platform and Dataset}

Experiments are conducted on an Imeta-Y1 robot equipped with three RGB cameras and DM tactile sensors. We consider three contact-rich manipulation tasks: \textbf{chip grasping}, which involves fragile objects with varying sizes and thicknesses; \textbf{fruit grasping}, which involves objects with diverse geometries and compliance; and \textbf{USB insertion}, which requires precise contact reasoning under visual occlusion. For each task, we collect 200 training trajectories and 40 test trajectories. All trajectories are recorded at 30\,Hz and downsampled to 6\,fps for world model training. Contact and slip labels are obtained from the official sensor API\@. Contact states are approximately balanced, whereas slip events account for only 0.93\% of the data, motivating the use of focal loss in Eq.~\eqref{eq:loss_slip}.

\subsubsection{Implementation Details}

FeelWorld adopts V-JEPA 2~\cite{assran2025v} as the visual backbone and processes two RGB views at a resolution of $224\times224$. Features from the two views are independently encoded and concatenated along the sequence dimension before being passed to the dynamics predictor. For a fair comparison, DINO-WM and JEPA-WM are extended to the same multi-view setting using an identical encoding strategy. The 3D tactile point clouds are encoded by a frozen FG-CLTP encoder $E_\tau$~\cite{ma2026fg} without fine-tuning. The dynamics predictor is a spatio-temporal transformer conditioned on 7-dimensional robot actions, while contact state is predicted by a two-layer MLP head and slip state by a causal 1D convolutional head. DINO-WM and JEPA-WM use their official decoders; since V-JEPA 2 provides none, we train a ViT-based decoder for V-JEPA 2, FeelWorld, and its ablations for fair visual comparison and visualization. The pretrained FG-CLTP decoder reconstructs tactile deformation maps. All decoders are used only for evaluation and visualization, not training or planning.

The model is trained on 8 H100 GPUs with a batch size of 64. We set $\lambda_\tau{=}0.3$, $\lambda_c{=}0.03$, $\lambda_s{=}0.03$, $\lambda_{\mathrm{roll}}{=}0.5$, $R{=}4$, and $\sigma{=}0.007$. During both training and evaluation, we use a 9-frame context window, corresponding to 1.5~s of observation history at 6~fps. Each zero-shot planning episode spans 50 control steps. At every replanning cycle, CEM samples $N{=}400$ candidate action sequences over a planning horizon of $H_p{=}6$, retains the top 8 candidates as elites, and performs 8 optimization iterations. The robot executes the first two actions before replanning. Planning performance is evaluated over 40 real-robot trials for each task.
\subsection{Visual Prediction Evaluation}
\label{subsec:visual_prediction_eval}

\begin{table*}[t]
\centering
\footnotesize
\caption{10-step rollout quality across visual baselines and FeelWorld ablations.}
\label{tab:results}
\resizebox{\textwidth}{!}{
\begin{tabular}{lccc|ccccc|ccc}
\hline
                           & \multicolumn{3}{c|}{Components} & \multicolumn{5}{c|}{Visual Prediction} & \multicolumn{3}{c}{Tactile Prediction} \\ \cline{2-12}
Method                     & Tac. & Hier. & Gate & PSNR$\uparrow$ & SSIM$\uparrow$ & LPIPS$\downarrow$ & FID$\downarrow$ & FVD$\downarrow$ & Chamfer$\downarrow$ & Contact F1$\uparrow$ & Slip F1$\uparrow$ \\ \hline
DINO-WM~\cite{zhou2025dino}                       & \xmark & \xmark & \xmark & 23.22 & 0.770 & 0.138 & 229.4 & 843.3 & -- & --  & -- \\
JEPA-WM~\cite{assran2023self}                    & \xmark & \xmark & \xmark & 23.78 & 0.781 & 0.132 & 204.3 & 792.1 & -- & --  & -- \\
V-JEPA 2 (visual-only)~\cite{assran2025v} & \xmark & \xmark & \xmark & 24.37 & 0.819 & 0.084 & 106.7 & 473.5 & -- & -- & -- \\
\rowcolor[HTML]{DAEFF2}FeelWorld (w/o hierarchical)          & \cmark & \xmark & \xmark & 26.71 & 0.853 & 0.071 & 79.14 & 326.4  & 0.072 & -- & -- \\
\rowcolor[HTML]{DAEFF2}FeelWorld (w/o contact gate)         & \cmark & \cmark & \xmark & 27.52 & 0.876 & 0.067 & 78.03 & 307.2  & 0.067 & 0.973 & 0.803 \\
\rowcolor[HTML]{DAEFF2} \textbf{FeelWorld (Ours)}          & \cmark & \cmark & \cmark & \textbf{28.89} & \textbf{0.884} & \textbf{0.058} & \textbf{76.45} & \textbf{289.9}  & \textbf{0.063} & \textbf{0.981} & \textbf{0.834} \\ \hline
\end{tabular}}
\end{table*}
\begin{figure*}[!t]
\centering
\includegraphics[width=\textwidth]{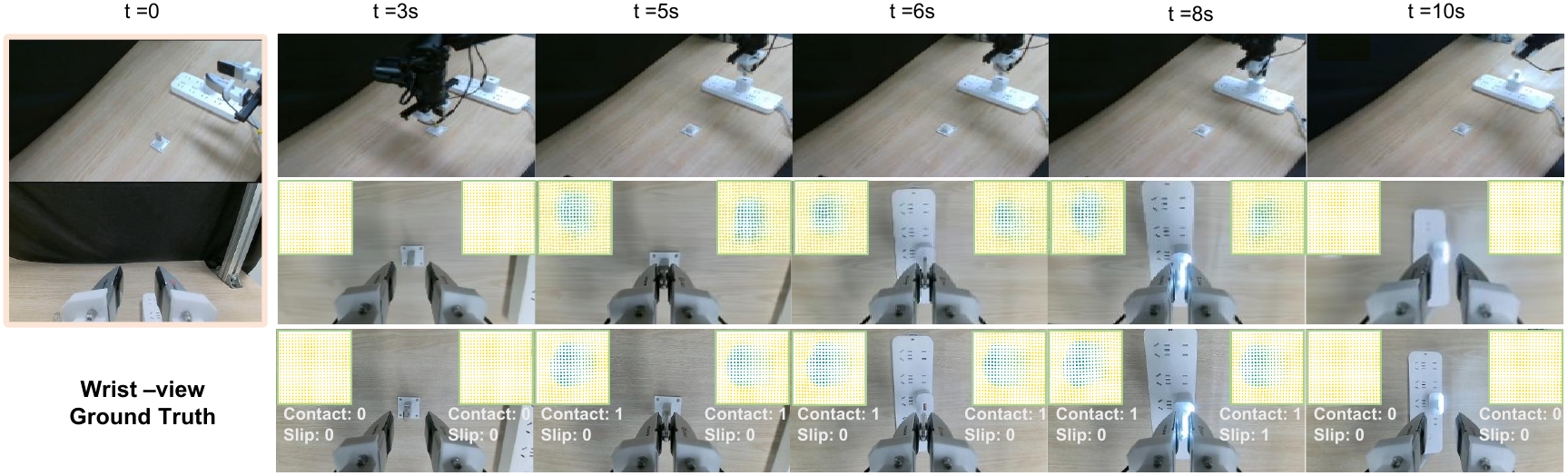}
\caption{\textbf{Qualitative visualizations of FeelWorld.} The model predicts future visual observations together with three tactile states: contact state, 3D tactile latent, and slip state. These tactile states change according to the interaction context and robot action.}
\label{fig:vis}
\end{figure*}
We evaluate the visual prediction quality of FeelWorld against several baselines. Table~\ref{tab:results} shows that the full model performs best across all five visual metrics. It is worth noting that approximately 30\% of the trajectories in our dataset are failed episodes, including unsuccessful grasps, object slippage, and failed USB insertions caused by subtle differences. Evaluating these trajectories provides a rigorous assessment of long-horizon prediction accuracy and helps determine whether the model captures fine-grained physical dynamics rather than overfitting to successful demonstrations. Compared with the visual-only V-JEPA 2 baseline, FeelWorld consistently improves visual quality and temporal consistency, increasing PSNR from 24.37 to 28.89 and SSIM from 0.819 to 0.884, while reducing LPIPS by 31.0\% and FVD by 38.8\%. The ablation results show that the basic naive visuo-tactile variant, which retains context noise injection and the rollout loss, already outperforms the visual-only model, reducing LPIPS from 0.084 to 0.071, as tactile observations provide complementary information for distinguishing interaction states and predicting the next frame around contact transitions. Explicit hierarchical tactile modeling further reduces LPIPS to 0.067, while asymmetric contact-gated attention achieves the best result of 0.058, with consistent improvements in PSNR and FVD\@. In particular, the full model preserves the visual prediction quality of the vision-only world model before contact by suppressing irrelevant tactile interference, while incorporating tactile information during contact for joint visuo-tactile learning and prediction. This selective integration substantially improves prediction consistency and demonstrates the effectiveness of the proposed hierarchical tactile modeling and contact-gated fusion.

\begin{table}[t]
\centering
\footnotesize
\caption{Visual prediction around contact transitions. Contact-onset error measures the deviation from ground-truth onset time.}
\label{tab:phase_breakdown}
\begin{tabular}{lccc}
\toprule
Method & Phase & LPIPS$\downarrow$ & Onset Error (frames)$\downarrow$ \\
\midrule
\multirow{2}{*}{V-JEPA 2} & Free space & 0.089 & -- \\
 & Contact transition & 0.127 & 1.17 \\
\midrule
\multirow{2}{*}{FeelWorld} & Free space & 0.046 & -- \\
 & Contact transition & 0.055 & 0.44 \\
\bottomrule
\end{tabular}
\end{table}

We then conduct a phase-wise analysis to evaluate prediction performance across different interaction stages. As shown in Table~\ref{tab:phase_breakdown}, visual prediction remains relatively reliable in free space but degrades substantially around contact transitions, where the object may slip, become stuck, or move together with the robot. The LPIPS of V-JEPA 2 increases by 42.7\%, from 0.089 to 0.127, indicating that vision alone cannot reliably infer grasp stability or impending slip. In contrast, the LPIPS of FeelWorld increases only from 0.046 to 0.055 and remains 56.7\% lower than that of V-JEPA 2 during contact transitions, demonstrating that tactile modeling reduces contact-induced ambiguity and improves prediction consistency. To further quantify the ambiguity of visual observations near contact, we freeze the V-JEPA 2 visual backbone and train only a lightweight contact-prediction head. The resulting contact-onset error is 1.17 frames, compared with only 0.44 frames when tactile features from FeelWorld are used, further highlighting the importance of tactile information for accurately identifying contact transitions.
\begin{figure}[!t]
\centering
\includegraphics[width=\columnwidth]{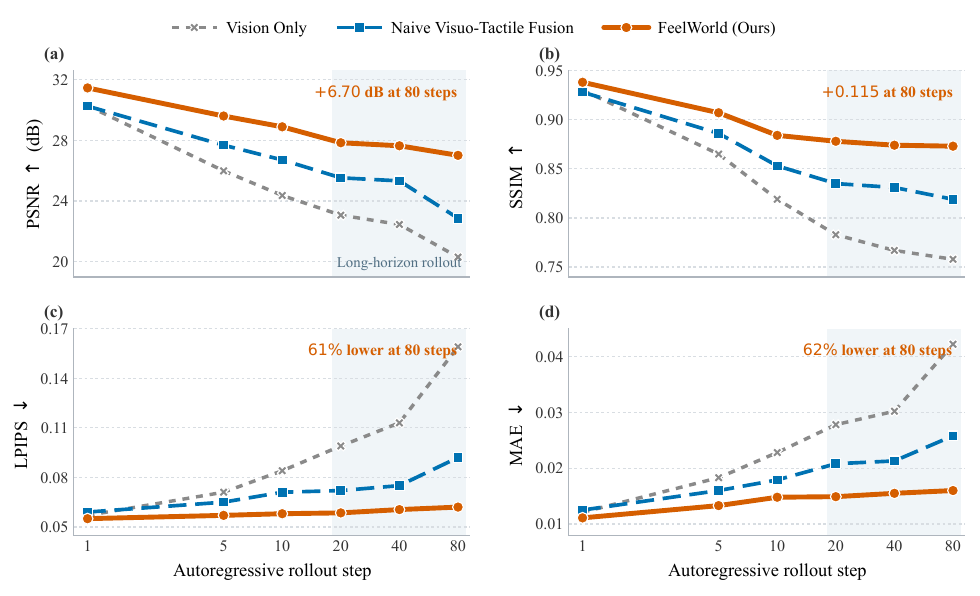}
\caption{\textbf{Long-horizon autoregressive visual prediction.} At step 80, FeelWorld improves PSNR and SSIM over Vision Only by 6.70\,dB and 0.115, while reducing LPIPS and MAE by 61\% and 62\%. The shaded region marks the long-horizon regime.}
\label{fig:rollout_robustness}
\end{figure}

We further evaluate the long-horizon rollout performance of FeelWorld through autoregressive prediction over 1, 5, 10, 20, 40, and 80 steps, where 80 steps correspond to approximately 13.3\,s. V-JEPA 2 corresponds to Vision Only, while FeelWorld (w/o hierarchical) corresponds to Naive Visuo-Tactile Fusion.
The experimental results show that the visual-only model exhibits substantial drift, which becomes increasingly pronounced as the rollout horizon increases, as shown in Fig.~\ref{fig:rollout_robustness}. Naive visuo-tactile fusion also drifts over long rollouts because unfiltered tactile signals perturb the visual prediction stream during non-contact phases. By suppressing this pre-contact interference, FeelWorld consistently achieves the best prediction performance, indicating that contact-gated attention and joint visuo-tactile modeling enable the model to capture fine-grained physical dynamics while supporting stable long-horizon rollouts. FeelWorld's LPIPS increases from 0.055 at step 1 to 0.062 at step 80 and remains 61\% lower than the visual-only baseline at the longest horizon.

\subsection{Tactile State Prediction}

We further evaluate the prediction accuracy of the three tactile states. As shown in Table~\ref{tab:state_prediction}, FeelWorld predicts contact with 98.6\% accuracy and a contact-onset error of only 0.44 frames. The tactile point cloud prediction error reported in Table~\ref{tab:results} is also low, indicating accurate modeling of local contact deformation. Slip metrics are computed over all test frames. Because slip-positive frames constitute only 0.93\% of the data, the dominant non-slip class yields a high raw accuracy of 99.7\%; F1, precision, and recall are therefore more informative. FeelWorld achieves 83.4\% F1, 81.7\% precision, and 85.2\% recall, demonstrating reliable detection of these rare but critical events. These results show that the fused tactile latent $\hat{z}^\tau$ effectively captures contact state, force-related information, and slip, providing essential physical grounding for visuo-tactile fusion.

\begin{figure}[!t]
\centering
\includegraphics[width=\columnwidth]{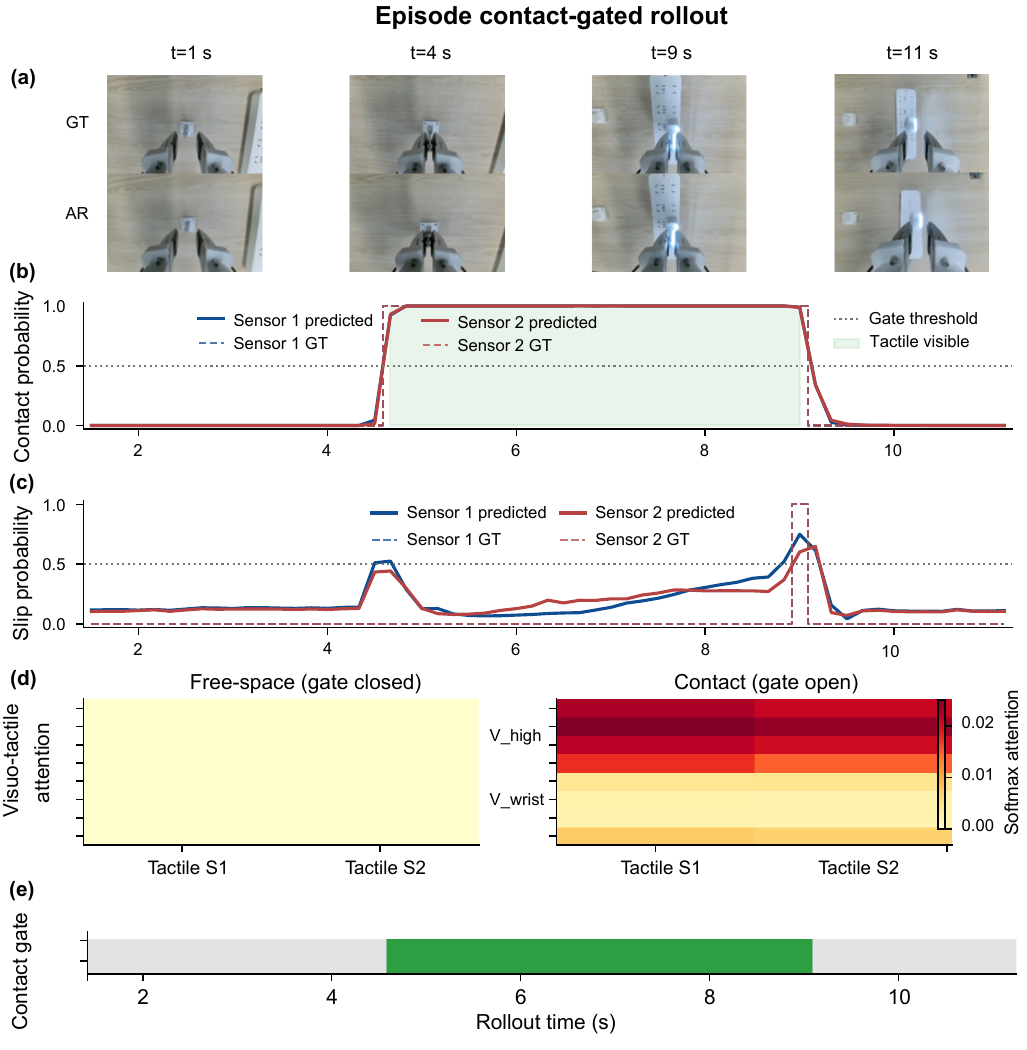}
\caption{\textbf{Visualization of contact-gated prediction during a USB insertion episode.} \textbf{(a)} Multi-stage visual rollout. \textbf{(b)} Ground-truth and predicted contact states. \textbf{(c)} Ground-truth and predicted slip states. \textbf{(d)} Gated visuo-tactile cross-attention for the external and wrist camera views. \textbf{(e)} Hard contact-gate activation over time.}
\label{fig:case_study}
\end{figure}

Fig.~\ref{fig:case_study} qualitatively illustrates the gating behavior during a USB insertion episode. Panel (a) shows the multi-stage visual rollout, while panels (b) and (c) compare the ground-truth and predicted contact and slip states. The model correctly identifies the slip risk during USB removal and the mild slip near the end of insertion. Panel (d) visualizes the gated asymmetric attention. Before contact, visuo-tactile cross-attention is suppressed during free-space motion; after contact, tactile information contributes more strongly to the external-view prediction than to the wrist-view prediction, likely because the external camera provides less local interaction detail. Panel (e) shows that the hard gate opens only during contact, demonstrating that tactile information is selectively incorporated into visual dynamics prediction during physical interaction.

\begin{table}[t]
\centering
\scriptsize
\caption{Contact and slip state prediction over all test frames. Onset and offset errors are in frames at 6\,fps.}
\label{tab:state_prediction}
\begin{tabular}{lcccccc}
\toprule
\multirow{2}{*}{State} & \multicolumn{4}{c}{Classification (\%)} & \multicolumn{2}{c}{Onset/Offset Error (frames) $\downarrow$} \\
\cmidrule(lr){2-5}\cmidrule(l){6-7}
 & Acc.$\uparrow$ & F1$\uparrow$ & Prec.$\uparrow$ & Rec.$\uparrow$ & Onset & Offset \\
\midrule
Contact & 98.6 & 98.1 & 98.5 & 97.8 & 0.44 & 0.69 \\
Slip & 99.7 & 83.4 & 81.7 & 85.2 & 0.56 & -- \\
\bottomrule
\end{tabular}
\end{table}

\begin{figure}[!t]
\centering
\includegraphics[width=\columnwidth]{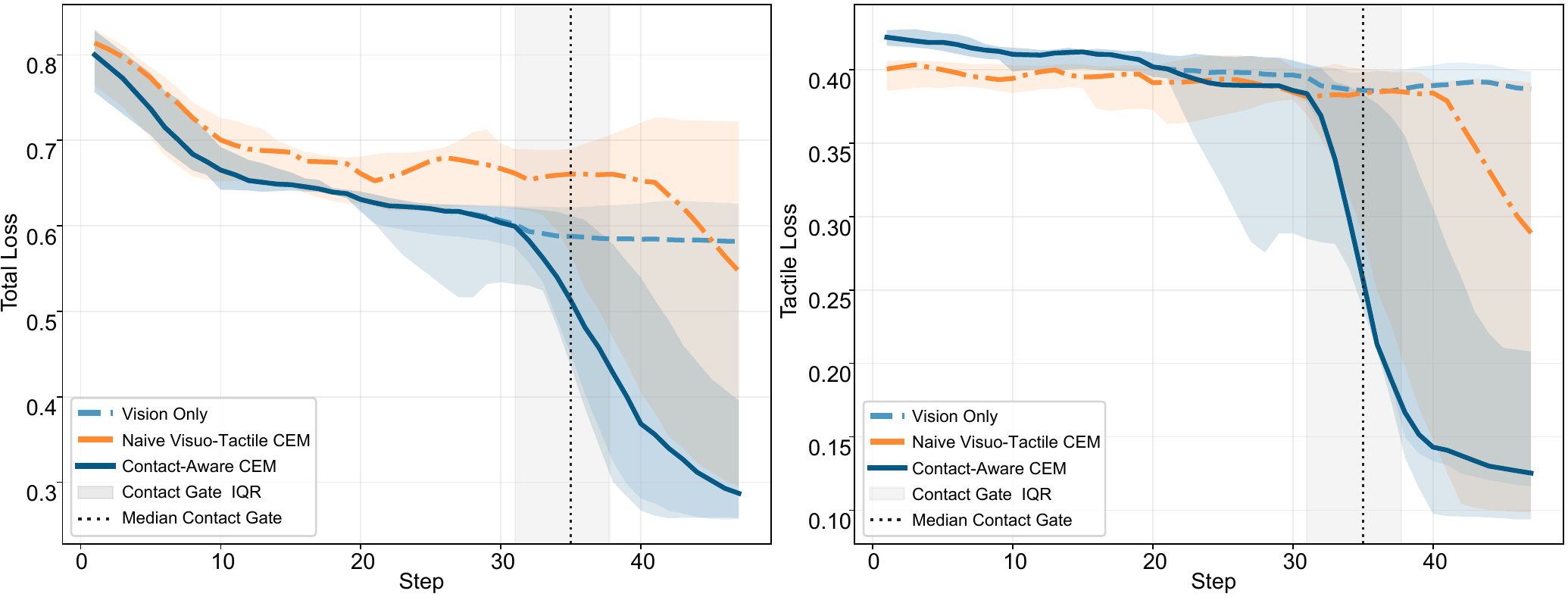}
\caption{\textbf{CEM planning curves over a 50-step execution.} The plots show total loss (left) and tactile loss (right). At each replanning cycle, CEM optimizes a 6-step imagined horizon and executes the first two actions. The dotted line and gray band mark the median and interquartile range of contact-gate activation.}
\label{fig:cem_convergence}
\end{figure}

\begin{table}[t]
\centering
\footnotesize
\caption{Zero-shot CEM planning success rates.}
\label{tab:planning}

\begin{tabular}{lccc}
\toprule
Planner & Chip & Fruit & USB \\
\midrule
Visual-only CEM (w/o Tactile) & 40.0 & 70.0 & 37.5 \\
Naive Visuo-Tactile CEM & 47.5 & 75.0 & 50.0 \\
\rowcolor[HTML]{DAEFF2} \textbf{Contact-aware CEM} & \textbf{82.5} & \textbf{87.5} & \textbf{75.0} \\
\bottomrule
\end{tabular}
\end{table}
\subsection{Zero-Shot Planning Results}
We evaluate zero-shot planning by specifying a task subgoal and selecting actions through imagined rollouts, without training a task-specific policy. Each episode runs for 50 control steps. At every replanning cycle, CEM samples 400 candidate action sequences over a planning horizon of $H_p{=}6$ imagined steps, executes the first two actions of the optimal sequence, and then replans. We compare three methods. \textbf{V-JEPA 2} is the visual world model baseline and optimizes only the visual latent distance to the goal. \textbf{Naive Visuo-Tactile CEM} jointly optimizes visual and tactile objectives throughout the planning horizon. \textbf{Ours} adopts the contact-gated planning objective in Eq.~\eqref{eq:planning_objective}.

Figure~\ref{fig:cem_convergence} shows the convergence of the planning objectives, while Table~\ref{tab:planning} reports the results of 40 real-robot trials for each task. Our method achieves success rates of 82.5\%, 87.5\%, and 75.0\% on chip grasping, fruit grasping, and USB insertion, respectively. In comparison, \textbf{V-JEPA 2} achieves 40.0\%, 70.0\%, and 37.5\%, while \textbf{Naive Visuo-Tactile CEM} achieves 47.5\%, 75.0\%, and 50.0\%.
The results highlight the importance of activating tactile objectives at the appropriate interaction stage. Vision-only CEM can guide the robot toward a visually plausible goal configuration, but visual similarity alone does not ensure stable contact or successful insertion. Naive visuo-tactile CEM improves contact-phase optimization, but optimizing an unreachable tactile goal before contact interferes with free-space approach. In contrast, our contact-gated CEM relies on visual guidance before contact and activates joint visuo-tactile optimization only after contact is predicted. This decomposition avoids pre-contact interference while enabling accurate reasoning about contact geometry, grasp stability, and slip during physical interaction. Consequently, our method consistently achieves the highest success rate across all tasks and improves USB insertion from 37.5\% to 75.0\% under severe visual occlusion.

\section{Conclusion}

We introduce FeelWorld, a hierarchical visuo-tactile world model that jointly predicts visual latents and three tactile states. Hierarchical tactile modeling with explicit supervision enables fine-grained representation of contact dynamics. The proposed contact-gated asymmetric attention suppresses irrelevant tactile interference during free-space motion while enabling joint visuo-tactile prediction during contact. Compared with the visual-only baseline and naive visuo-tactile fusion, FeelWorld achieves superior 10-step prediction accuracy and long-horizon rollout robustness. It reduces 10-step LPIPS from 0.084 to 0.058. Over an 80-step autoregressive rollout, LPIPS increases to 0.062 and remains 61\% lower than that of the visual baseline. FeelWorld also achieves F1 scores of 98.1\% and 83.4\% for contact and slip prediction, respectively. Furthermore, contact-aware CEM planning achieves an average success rate of 81.7\%, exceeding the visual-only baseline by 32.5 percentage points. These results demonstrate that hierarchical tactile modeling and contact-gated asymmetric attention enable effective visuo-tactile fusion and contact-state understanding, providing a general framework for incorporating tactile sensing into world models.

Despite these results, CEM-based planning remains computationally expensive and is not suitable for high-frequency real-time control. Future work could integrate FeelWorld with a policy network to efficiently evaluate candidate rollouts and iteratively refine the model using real-world interaction data, thereby providing physically consistent guidance for policy learning and optimization in contact-rich manipulation tasks.
\bibliographystyle{IEEEtran}
\bibliography{example}

\end{document}